\documentclass{article}
\usepackage{ijcai22}

\usepackage{times}
\usepackage{soul}
\usepackage{url}
\usepackage[hidelinks]{hyperref}
\usepackage[utf8]{inputenc}
\usepackage[small]{caption}
\usepackage{graphicx}
\usepackage{amsmath}
\usepackage{amsthm}
\usepackage{booktabs}
\usepackage{algorithm}
\usepackage{algorithmic}
\usepackage{xcolor}
\usepackage{multirow}
\usepackage{multicol}
\usepackage{tablefootnote}
{\makeatletter\long\gdef\@gobble#1{}}

\usepackage{subfig}
\usepackage{graphicx}

\usepackage{makecell}
\usepackage{bbm}
\usepackage{amsmath}
\usepackage{amssymb}

\def \submission {}
\ifx \submission \undefined
    \newcommand{\hw}[1]{{\color{teal}{\bf\sf [haohan: #1]}}}
    \newcommand{\oscar}[1]{{\color{green}{\bf\sf [oscar: #1]}}}
    \newcommand{\chonghan}[1]{{\color{olive}{\bf\sf [chonghan: #1]}}}
    \newcommand{\noel}[1]{{\color{orange}{\bf\sf [noel: #1]}}}
    \newcommand{\jq}[1]{{\color{blue}{\bf\sf [jq: #1]}}}
    \newcommand{\xl}[1]{{\color{orange} \bf \sf (XL: #1)}}
\else
    \newcommand{\hw}[1]{{\iffalse{#1}\fi}}
    \newcommand{\oscar}[1]{{\iffalse{#1}\fi}}
    \newcommand{\chonghan}[1]{{\iffalse{#1}\fi}}
    \newcommand{\noel}[1]{{\iffalse{#1}\fi}}
    \newcommand{\jq}[1]{{\iffalse{#1}\fi}}
    \newcommand{\xl}[1]{{\iffalse{#1}\fi}}
\fi

\title{Bear the Query in Mind: Visual Grounding with Query-conditioned Convolution}

\author{
Chonghan Chen$^1$\footnote{Equal Contribution}\and
Qi Jiang$^1$\footnotemark[1]\and
Chih-Hao Wang$^1$\footnotemark[1]\and
Noel Chen$^1$\footnotemark[1]\and 
Haohan Wang$^{1,2}$\and \\ 
Xiang Li$^1$\and 
Bhiksha Raj$^1$ \\
\affiliations
$^{1,2}$School of Computer Science, Carnegie Mellon University \\
$^{2}$ School of Information Science, University of Illinois Urbana-Champaign
\emails
\{chonghac, qij, chihaow, yunhsua3, haohanw, xl6\}@andrew.cmu.edu,\\
bhiksha@cs.cmu.edu
}

\begin{document}

\maketitle

\begin{abstract}
Visual grounding is a task that aims to locate a target object according to a natural language expression.
As a multi-modal task, feature interaction between textual and visual inputs is vital. 
However, previous solutions mainly handle each modality independently before fusing them together, which does not take full advantage of relevant textual information while extracting visual features.
To better leverage the textual-visual relationship in visual grounding, we propose a \textbf{Q}uery-conditioned \textbf{C}onvolution \textbf{M}odule (QCM) that extracts query-aware visual features by incorporating query information into the generation of convolutional kernels.
With our proposed QCM, the downstream fusion module receives visual features that are more discriminative and focused on the desired object described in the expression, leading to more accurate predictions. 
Extensive experiments on three popular visual grounding datasets demonstrate that our method achieves state-of-the-art performance. In addition, the query-aware visual features are informative enough to achieve comparable performance to the latest methods when directly used for prediction without further multi-modal fusion.


\end{abstract}

\section{Introduction}

Visual Grounding, also known as Referring Expression Comprehension \cite{refcocog,refcocoandplus}, is a task of locating a target object in an image according to a natural language query. It requires the machine to understand both textual and visual information, as well as the relationship between them.
Pioneer works \cite{hu2017modeling,zhang2018grounding} of visual grounding mainly utilize a two-stage propose-and-rank scheme, and then the focus has shifted to one-stage methods \cite{Liao_2020_CVPR,yang2020improving} that directly predict the bounding box coordinates. 
With the success of transformer-based models \cite{vaswani2017attention} in recent years, a novel transformer-based one-stage paradigm for visual grounding emerges. 


\begin{figure}[tbp]
    \centering
    \includegraphics[width=\linewidth]{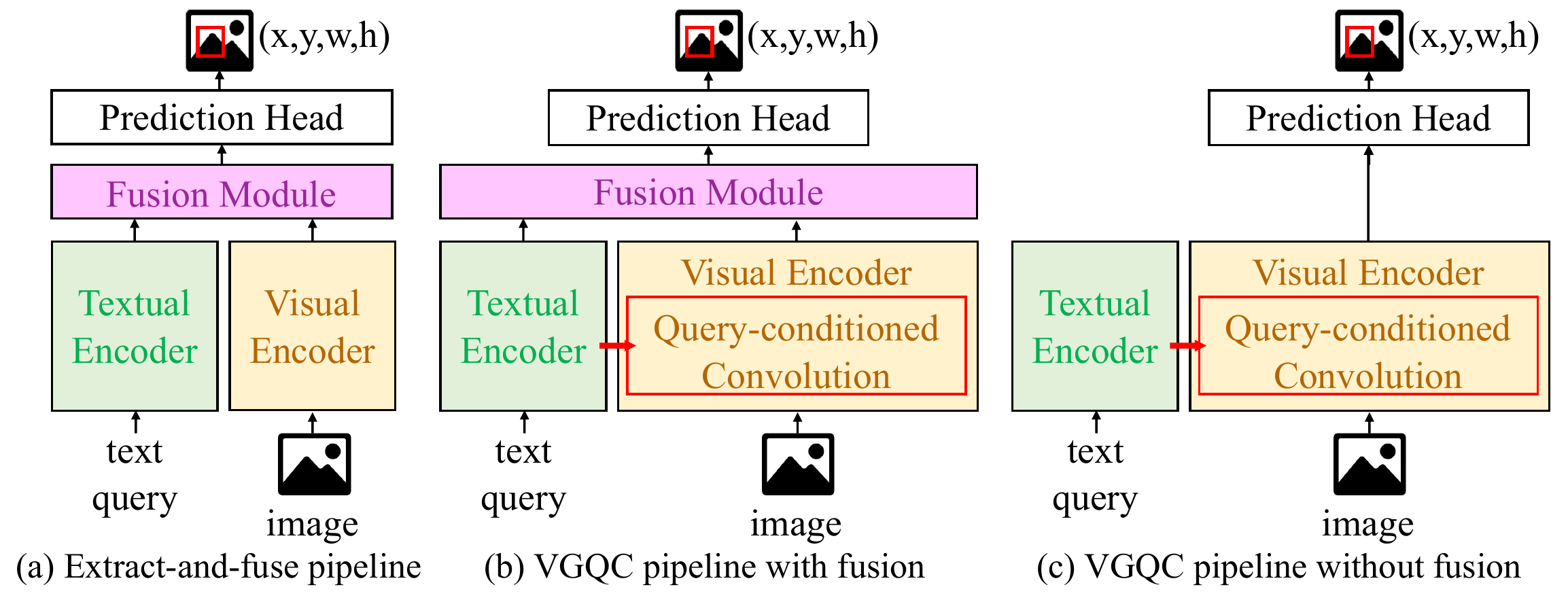}
    \caption{Previous extract-and-fuse pipeline (a) and our VGQC pipeline (b) and (c). In (a), textual and visual features are extracted independently and fused together in the following fusion module. 
    (b) uses a visual encoder with query-conditioned convolution that takes in both the image and query token to produce query-aware visual features. 
    (c) removes the fusion module as query-aware visual features are informative enough to be used directly for prediction.}
    \label{fig:vis-pipleine-comparison}
\end{figure}

As shown in Figure~\ref{fig:vis-pipleine-comparison}(a), most previous methods~\cite{chen2018realtime,Liao_2020_CVPR,du2021visual,yang2020improving,deng2021transvg} follow the extract-and-fuse pipeline that extracts visual and textual features independently using two isolated encoders followed by an additional fusion module to enable multi-modal feature interaction.
As the visual encoder is unaware of the text query during feature extraction, it only encodes visual features based on intra-image information, neglecting the potential correspondence between text query and target image.
Consequently, the extracted visual features may not convey the required information described by the text query.
In this case, the visual encoder is likely to pass redundant or even misleading visual features to the following fusion module and thus compromise the performance. Intuitively, the extract-and-fuse pipeline is similar to asking humans to first memorize an image and then recall it while locating an object given a natural language description. The unnecessary memorization makes the problem difficult and violates the nature of human behaviors which typically locates the object while understanding the image.

To address this problem, we follow the intuition of human behavior that knowing what to look for before looking at the image will make object localization problems easier. For a neural network, this is equivalent to asking the visual encoder to understand the text query and extract relevant visual features guided by the text query. 
Dynamic convolution \cite{landi2019embodied,chen2020dynamic}, which typically controls feature generation with internal or external signals, \textit{de facto} fits the need for text-aware visual feature extraction.
Inspired by a previous dynamic convolution method \cite{chen2020dynamic}, we propose a query-conditioned convolution module (QCM) that dynamically constructs convolutional kernels according to input queries. In this way, the output visual features are more discriminative for each query-image pair and lead to more accurate predictions. 
To enable query-aware feature extraction, we replace the vanilla convolution in the visual encoder with QCM in a multi-scale fashion. We term our method as VGQC, a \textbf{V}isual \textbf{G}rounding pipeline with \textbf{Q}uery-\textbf{C}onditioned visual encoder.
Different from other fusion methods \cite{Joze2020MMTMMT,nagrani2021attention} that enable unbiased multi-modal interaction bi-directionally, our method solely imports textual information into the visual representation, and it happens during the process of extracting visual features instead of the post-processing of extracted static visual features. 
In addition, query-aware visual features generated by QCM are informative enough to be directly used for prediction without the requirement for additional multi-modal fusion modules used in previous methods~\cite{deng2021transvg,du2021visual}, resulting in a simpler VGQC pipeline. 
The experiment results show that VGQC w/ fusion achieves state-of-the-art performance on nearly all testing datasets, and VGQC w/o fusion increases the inference speed while achieving comparable performance as the latest methods. Then, extensive analysis and visualization illustrate the effectiveness of QCM and query-aware visual features.

In summary, the contribution of our work is three-fold: 
\begin{enumerate}
    \item We present a novel query-conditioned convolution module (QCM) that extracts query-aware visual features and can be flexibly integrated into visual encoders. We also introduce a visual grounding pipeline VGQC which utilizes QCM to address the problem in previous extract-and-fuse methods. 
    \item  Our experiments show that VGQC w/ fusion achieves state-of-the-art performance on RefCOCO \cite{refcocoandplus}, RefCOCO+ \cite{refcocoandplus}, and RefCOCOg \cite{refcocog}, and VGQC w/o fusion achieves comparable performance with faster inference speed and a simpler model structure.
    \item We conduct extensive analysis with visualizations to demonstrate the diversity of candidate kernels as well as the effectiveness of encoding query information into QCM's attention weights and aggregated kernels, which enable such kernels to extract query-aware visual features.
\end{enumerate}

\section{Related Work}
\subsection{Visual Grounding}


\paragraph{Two-stage methods.}
Two-stage methods, also known as propose-and-rank methods, generate image region candidates and then rank them using the information from text queries to select the best matching one. The candidates are usually generated using a pre-trained object detector \cite{wang2019neighbourhood} or unsupervised methods \cite{yu2018mattnet}. Some early methods utilize modular networks to compute matching scores \cite{yu2018mattnet}, and recently many two-stage methods focus on modeling the inter-object relationships using graph representations. \cite{yang2019dynamic,wang2019neighbourhood,hong2019learning,Liu_Wan_Zhu_He_2020}.

\paragraph{One-stage methods.}
One-stage methods directly predict the bounding boxes and most of them adopt the extract-and-fuse pipeline. RCCF~\cite{Liao_2020_CVPR} uses a cross-modality correlation filtering on visual and textual features, and 
ReSC \cite{yang2020improving} proposes a recursive sub-query construction framework to handle complex queries. 
With the success and popularity of transformers~\cite{vaswani2017attention}, recent one-stage methods have adopted transformer-based fusion modules and have achieved state-of-the-art performance~\cite{du2021visual,deng2021transvg}. 

\subsection{Multi-modal Feature Fusion}
We also review several multi-modal feature fusion methods. 
\cite{landi2019embodied} constructs dynamic convolution filters from text and post-processes the output feature maps with the filters. Compared to dynamic convolution \cite{chen2020dynamic} which composes convolution kernels with attention weights derived from the previous feature map, it directly generates convolution kernels from text.
Another convolution-based method MMTM \cite{Joze2020MMTMMT} 
learns a joint representation and generates an excitation signal for each modality.
In transformer-based fusion methods, \cite{nagrani2021attention} proposes fusion bottlenecks in transformer encoders, while \cite{Zhao_2021_ICCV} appends cross-modal attention after self-attention in their fusion transformer encoders.

    


\begin{figure}[t]
    \centering
    
    \includegraphics[width=1\linewidth]{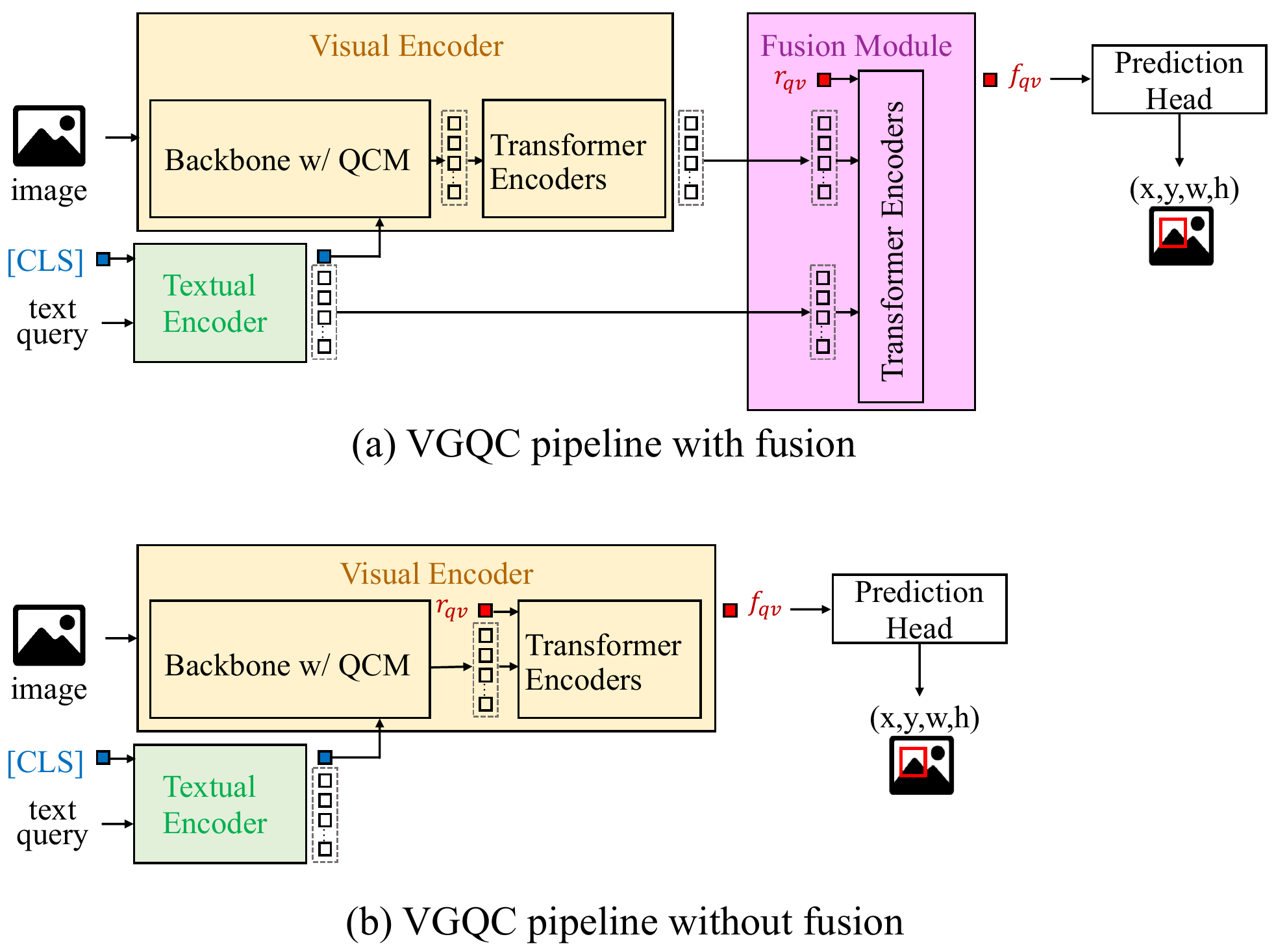}
    \caption{Two variants of our query-conditioned visual grounding (VGQC) pipelines. (a) combines query-aware visual features and query features with a fusion module, while (b) removes the fusion module and passes query-aware visual features directly into the prediction head.
    }
    \label{fig:framework}

\end{figure}

\section{Methodology}


We leverage the transformer-based network to tackle the visual grounding problem. Unlike previous methods \cite{chen2018realtime,yang2020improving,deng2021transvg,chen-etal-2021-multimodal} that extract visual and textual features separately, we encourage query-aware visual feature extraction in the visual encoder. With query-aware visual features, the downstream multi-modal fusion \cite{yang2020improving,deng2021transvg} becomes optional. Figure~\ref{fig:framework} shows the overall pipeline of our proposed method. In particular, our VGQC network can be boiled down to four parts: a textual encoder, a visual encoder, an optional fusion module, and a prediction head.

\subsection{Textual Encoder}

The textual encoder extracts textual features from input expressions at both sentence and word levels. Following previous works \cite{deng2021transvg,yang2020improving}, we use BERT \cite{devlin2019bert} as our textual encoder. The input to the textual encoder is the concatenation of a [CLS] token, word tokens, and a [SEP] token. The output of the textual encoder is the query token $f_{query} \in \mathbb{R}^{C_q\times 1}$ encoding the sentence level information of the given query and a set of word tokens of the input query $f_q \in \mathbb{R}^{C_q\times{N_q}}$ where $N_q$ is the number of word tokens, and $C_q$ is set to 768 for each token.
 
\begin{figure}[t]
  \centering
  \includegraphics[width=0.9\linewidth]{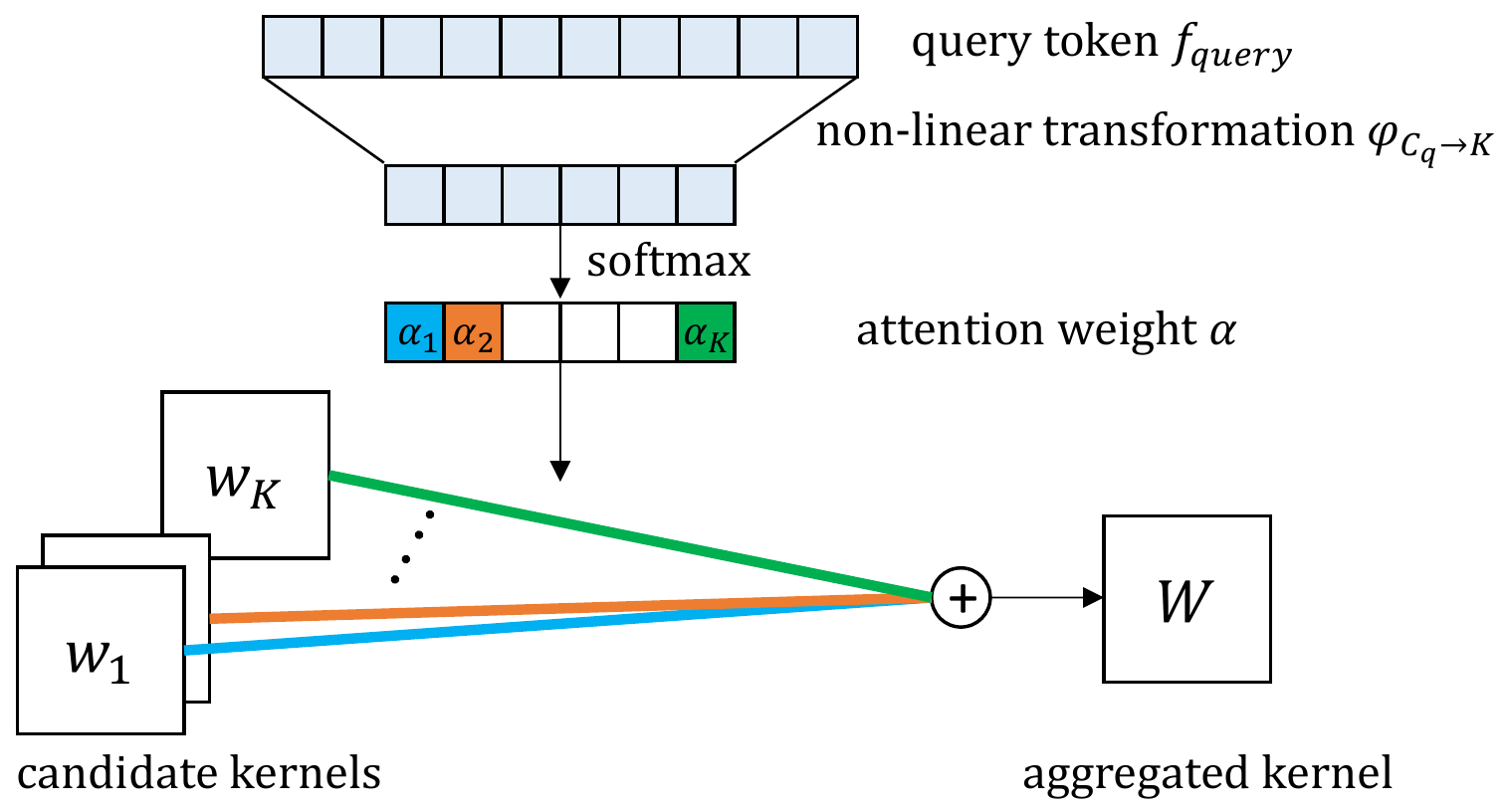}
  \caption{The structure of a query-conditioned convolution module (QCM). There are K candidate kernels $\{\boldsymbol{w}_1, \boldsymbol{w}_2,..., \boldsymbol{w}_K\}$. An attention weight $\boldsymbol{\alpha}$ is derived from the query token $f_{query}$, and the aggregated kernel $\boldsymbol{W}$ is a weighted sum of the candidate kernels.}
  \label{fig:aggregated-kernel_block}
\end{figure}
\subsection{Visual Encoder}
\label{sec:visual-encoder}

To enable query-aware visual feature extraction, we propose a query-conditioned convolution module (QCM) and replace parts of the vanilla convolution in the visual encoder with it in a multi-scale manner.

\paragraph{Query-conditioned convolution.}
A QCM block takes in the query token $f_{query}$ to compose a convolution kernel to extract query-relevant features from an input feature map. As shown in Figure~\ref{fig:aggregated-kernel_block}, in each QCM block, there are $K$ learnable candidate convolution kernels $\{\boldsymbol{w}_1, \boldsymbol{w}_2,..., \boldsymbol{w}_K\}$ where $\boldsymbol{w}_i \in \mathbb{R}^{C_{in}\times{C_{out}}\times{N}\times{N}}$, and each candidate kernel represents a particular set of features to be extracted. As text queries with different semantic meanings may require different combinations of these sets of features, the candidate kernels are aggregated according to an attention weight $\boldsymbol{\alpha}=[\alpha_1, ..., \alpha_K]  \in \mathbb{R}^{K\times{1}}$ computed from the query token $f_{query}$. The attention weight $\boldsymbol{\alpha}$ is calculated by projecting $f_{query}$ into a $K$-dimensional vector using a non-linear projection and a softmax layer:
\begin{equation}
   \boldsymbol{\alpha} = \mathrm{Softmax}(\varphi_{C_q \mapsto K} (f_{query}))
\end{equation}
where $\varphi_{C_q \mapsto K}(\cdot)$ is a non-linear transformation function.
The aggregated convolution kernel $\boldsymbol{W} \in \mathbb{R}^{C_{in}\times{C_{out}}\times{N}\times{N}}$ is a weighted sum of the candidate kernels computed as:
\begin{equation}
    \boldsymbol{W} = \sum^{K}_{i = 1}\alpha_i\boldsymbol{w}_i
\end{equation}
By incorporating query information into the generation of convolutional kernels, QCM blocks can extract distinct features according to different text queries to produce visual features that are more relevant to target objects.

\paragraph{Visual encoder.}

The visual encoder consists of a backbone equipped with QCM followed by a stack of transformer encoders to extract query-aware visual features. 
Specifically, to consider multi-scale features, we replace the first vanilla convolution in the backbone in $[\frac{1}{4}, \frac{1}{8}, \frac{1}{16}, \frac{1}{32}]$ resolution stages with our query-conditioned convolution. Let us denote the extracted visual features from the backbone as $f_v^{\prime}$ with channel size of $C_v^\prime$. A sine spatial positional encoding \cite{carion2020endtoend} is added to the projected feature map before feeding it to the transformer encoders.
\begin{equation}
    f_v^{\prime\prime}=\mathrm{Flatten}(\varphi_{C_v^\prime\mapsto C_v}(f_v^\prime)+P)
\end{equation}
where $f_v^{\prime\prime}$ is the input to the transformer encoders, $P$ is the positional encoding, and $\varphi_{C_v^\prime\mapsto C_v}(\cdot)$ is a dimensional reduction operation.
A stack of transformer encoders is further leveraged to model the non-local correspondence in the extracted feature map. The final output of the visual encoder $f_v \in \mathbb{R}^{C_v\times N_v}$ is a set of query-aware visual tokens, where $C_v$ is the feature dimension and $N_v$ is the number of the visual tokens. 

\subsection{Multi-modal Fusion}

After deriving the textual tokens $f_q$ and query-aware visual tokens $f_v$ from the encoders, they are further passed into a fusion module to generate a grounding representation. Following previous methods \cite{du2021visual,deng2021transvg,nagrani2021attention}, we utilize transformer encoders to conduct multi-modal interaction. As shown in Figure~\ref{fig:framework}(a), the transformer encoders take both $f_v$ and $f_q$ as inputs. In particular, the textual and visual tokens are first projected into the same dimension $C_r$ separately. Let us denote the projected visual tokens as $r_q \in \mathbb{R}^{C_r\times{N_q}}$ and textual tokens as $r_v\in \mathbb{R}^{C_r\times{N_v}}$. In addition, to force multi-modal interaction, we add a regression token $r_{qv} \in \mathbb{R}^{C_r\times{1}}$ into the transformer encoders. The final input to the multi-modal transformer encoders is the concatenation of the regression token, visual tokens and textual tokens $r_{qv}\oplus r_{v}\oplus r_{q}$. In the self-attention layers of the transformer encoders, the regression token $r_{qv}$ can learn from all the textual and visual tokens as well as the relationships between them. Therefore, its corresponding output after interaction $f_{qv}$ represents the global information encoded in the query-image pair. $f_{qv}$ is later passed into the prediction head as the grounding representation.

To show the power of our query-conditioned convolution, we remove the optional fusion module in the VGQC pipeline as shown in Figure~\ref{fig:framework}(b). After deriving the query-aware visual features $f_v^\prime$ from the backbone, a regression token $r_{qv} \in \mathbb{R}^{C_v\times{1}}$ is appended in front of $\varphi_{C_v^\prime\mapsto C_v}(f_v^\prime)$ with positional encoding $P$ before being passed into the transformer encoders. As $f_v^\prime$ already contains textual information, the regression token $r_{qv}$ can also learn both the textual and visual information. 
The corresponding output $f_{qv}$ is considered as the grounding representation and can be directly used in the prediction head.

\subsection{Prediction Head}
The final bounding box prediction is derived from the grounding representation $f_{qv}$ as:
\begin{equation}
    b = \varphi_{C_v\mapsto 4}(f_{qv})
\end{equation}
where $\varphi_{C_v\mapsto 4}(\cdot)$ is two fully connected layers with ReLU activatiton. $b$ is the coordinates of the predicted bounding box.

\subsection{Training Objective}
The training objective is the GIoU loss plus a smooth L1 loss. Let $A(S)$ denote the area covered by a bounding box $S$, and let $S_i$ and $\hat{S}_i$ be the ground-truth and predicted bounding boxes for sample $d_i$, respectively. Intersection-over-Union (IoU) ratio is calculated as:
\begin{equation}
    IoU_i = \frac{A(S_i \cap \hat{S}_i )}{A(S_i \cup \hat{S}_i )}
\end{equation}
In Generalized IoU ~\cite{rezatofighi2019generalized}, let $S_c$ denote the smallest convex object enclosing $S_i$ and $\hat{S}_i$, and GIoU is calculated as:
\begin{equation}
    GIoU_i = IoU_i - \frac{A(S_c \setminus (S_i \cup \hat{S}_i))}{A(S_c)}
\end{equation}
where $X \setminus Y$ means $X$ excluding $Y$.
GIoU loss is $L_{GIoU}=1-GIoU$. The smooth L1 loss is calculated between the ground-truth coordinates and the predicted coordinates normalized by the size of the image. The training objective is to minimize the overall loss $L$:
\begin{equation}
    L = L_{GIoU} + L_{smooth_{l1}}
\end{equation}
\section{Experiments}
\subsection{Datasets and Implementation Details}
\paragraph{Dataset.} We evaluate our models on three widely used datasets, RefCOCO~\cite{refcocoandplus}, RefCOCO+~\cite{refcocoandplus}, and RefCOCOg~\cite{refcocog}. All three of them share image data from COCO~\cite{coco} with additional referring expressions.
On average, RefCOCOg has longer expressions than RefCOCO and RefCOCO+ (8.4 v.s. 3.5 words per expression) and has more complicated expressions with more attributes. 
There are two types of data partitions for the RefCOCOg dataset, RefCOCOg-google~\cite{refcocog} and RefCOCOg-umd~\cite{nagaraja2016modeling}. We evaluate VGQC's performance on both of them.

\paragraph{Implementation details.}

The following is the best hyper-parameter setting that we have experimented with. We use a learning rate of $0.0002$, a batch size of 32, and a step scheduler that shrinks the learning rate by a factor of 0.1 every 60 epochs. We use AdamW optimizer \cite{loshchilov2017decoupled} with a $\mathrm{weight \,decay}=10^{-4}$ for all experiments. We choose ResNet-50 \cite{he2015deep} as  the visual backbone and apply QCM with $K=5$ on each stage. A pre-trained BERT-base \cite{devlin2019bert} is adopted as the textual encoder. We train our model for 90 epochs on RefCOCO and RefCOCOg, and 180 epochs for RefCOCO+. Data augmentation is adopted to obtain a strong baseline. Each image is padded and resized to $640\times 640$. Note that for RefCOCOg, we report the scores of the models trained on RefCOCO. More implementation details can be found in the supplementary materials.

\begin{table*}[t]
	\vspace{-0.05cm}
	\small
	\begin{center}
		\scalebox{0.9}[0.9]{
			\setlength
			\tabcolsep{8.4pt}
			\begin{tabular}{l | r r r | r r r | r r r | r}
				\toprule
				\multirow{2}{*}{Models} & \multicolumn{3}{c|}{RefCOCO} & \multicolumn{3}{c|}{RefCOCO+} & \multicolumn{3}{c|}{RefCOCOg} & Time \\ 
				
				& val & testA & testB & val & testA & testB & val-g & val-u & test-u & (ms)  \\
				\hline
				\textbf{\textit{Two-stage:}} & &  & & & & & & & &\\
				LGRANs~\cite{wang2019neighbourhood} & - & 76.60 & 66.40 & - & 64.00 & 53.40 & 61.78 & - & - & - \\
				DGA~\cite{yang2019dynamic} & - & 78.42 & 65.53 & - & 69.07 & 51.99 & - & - & 63.28 & - \\ 
				RvG-Tree~\cite{hong2019learning} & 75.06 & 78.61 & 69.85 & 63.51 & 67.45 & 56.66 & - & 66.95 & 66.51 & - \\
				NMTree~\cite{liu2019learning} & 76.41 & 81.21 & 70.09 & 66.46 & 72.02 & 57.52 & 64.62 & 65.87 & 66.44 & - \\
				\hline
				\textbf{\textit{One-stage:}} & &  & & & & & & & &\\
				RCCF~\cite{Liao_2020_CVPR} & - & 81.06 & 71.85 & - & 70.35 & 56.32 & -  & - & 65.73 & - \\
				ReSC-large~\cite{yang2020improving} & 77.63 & 80.45 & 72.30 & 63.59 & 68.36 & 56.81 & 63.12 & 67.30 & 67.20 & - \\
				VGTR~\cite{du2021visual} & 78.29 & 81.49 &  72.38 & 63.29 & 70.01 & 55.64 & 61.64 & 64.19 & 64.01 & - \\
				TransVG~\cite{deng2021transvg} & 80.18 & 82.53 & 75.25 & 65.47 & 70.09 & 57.20 & 66.23 & 66.87 & \textbf{67.94} & 30.99 \\
				\hline
				\textbf{\textit{Ours:}} & &  & & & & & & & & \\
				VGQC w/ fusion & \textbf{82.47}    & \textbf{85.02}   & \textbf{77.98}   & \textbf{69.17}  & \textbf{73.52} & \textbf{58.97}  & \textbf{68.72} & \textbf{67.95} & 67.89 & 31.74 \\
				VGQC w/o fusion & 80.01    & \textbf{83.02}   & 73.94   & \textbf{66.37}    & \textbf{73.11}    & 56.50   & 65.50 & 65.44 & 65.75 & \textbf{27.92}  \\ \hline
			\end{tabular}
		} 
	\end{center}
	\caption{Comparison of top-1 accuracy (\%) with previous methods on RefCOCO~\protect\cite{refcocoandplus}, RefCOCO+~\protect\cite{refcocoandplus} and RefCOCOg~\protect\cite{refcocog}. ``val-g" stands for RefCOCOg-google’s validation set, and ``val-u", ``test-u" stand for RefCOCOg-umd’s validation, test set, respectively. The results of TransVG~\protect\cite{deng2021transvg} are reproduced using its released source code. The column ``Time" shows the inference time of a single sample. We repeat the inference time measurement over the entire RefCOCO testA set on an NVIDIA T4 GPU and report the average.} 
	\label{tab:refcoco_results}
\end{table*}

\subsection{Quantitative Results}
In Table~\ref{tab:refcoco_results}, we compare our VGQC with two-stage and one-stage visual grounding models, among which TransVG is the previous state-of-the-art which doesn't exploit extra data. A prediction is considered correct if the IoU between the prediction and the ground-truth bounding boxes is larger than 0.5. We report the top-1 accuracy (\%) on RefCOCO, RefCOCO+, and RefCOCOg. The results show that our VGQC w/ fusion achieves state-of-the-art performance on all three datasets. On testA and testB of RefCOCO and RefCOCO+, our accuracy scores reach 85.02\%, 77.98\%, 73.52\%, and 58.97\% with an absolute improvement of 2.49\%, 2.73\%, 3.43\% and 1.77\%. On RefCOCOg, VGQC w/ fusion outperforms previous methods on the validation sets and achieves similar performance on the test set.
Even after removing the fusion module, VGQC w/o fusion also achieves comparable performance to previous methods on RefCOCO and RefCOCOg, and it outperforms TransVG on RefCOCO+ with less reference time. 

\subsection{Qualitative Analysis}
\begin{figure}[t]
  \centering
    
  \includegraphics[width=1\linewidth]{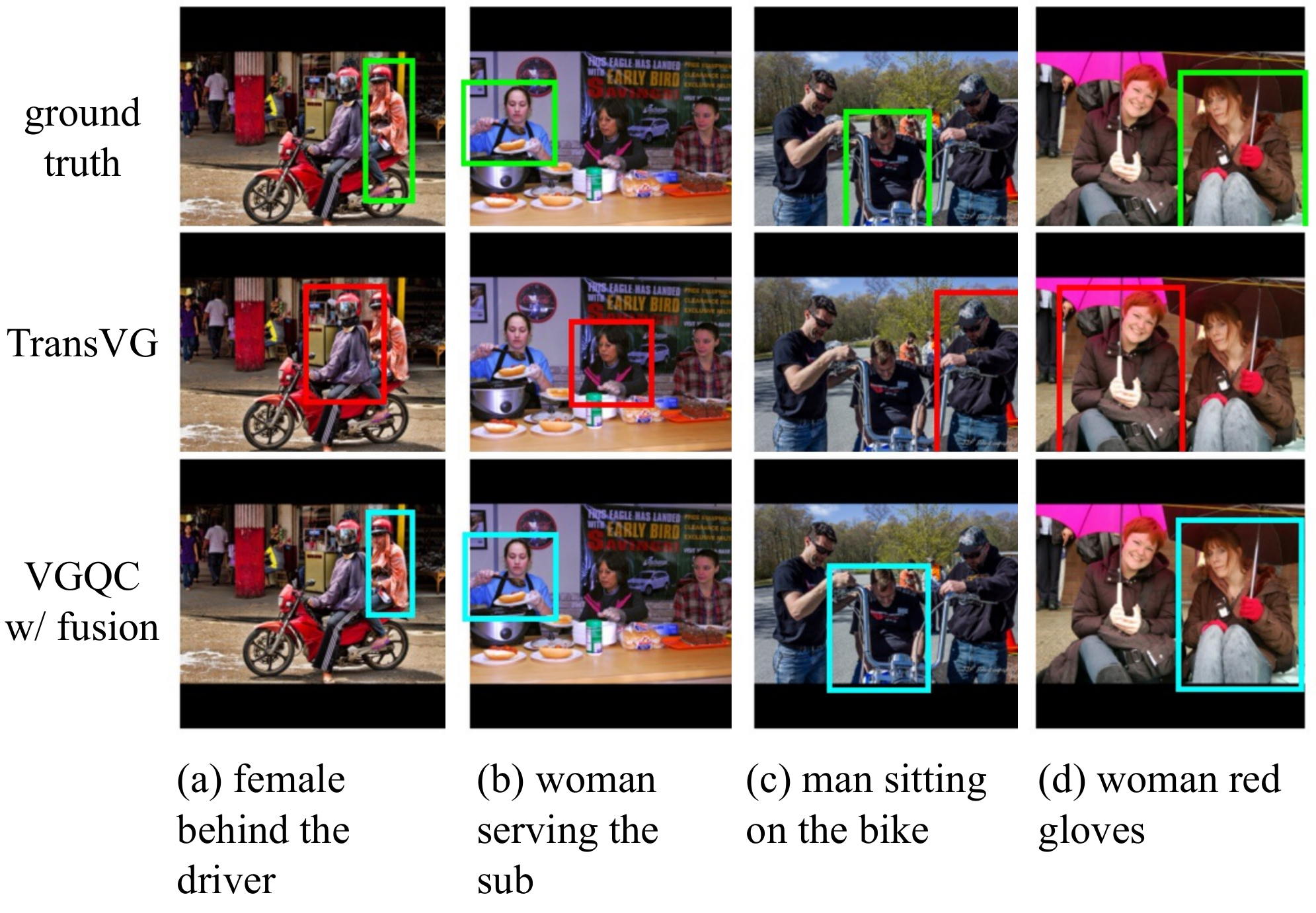}
  \caption{Qualitative results of VGQC w/ fusion on RefCOCO+ testA set. The \textcolor{green}{green} boxes are the ground truth labels, the \textcolor{red}{red} boxes are the predicted labels from TransVG, the \textcolor{cyan}{blue} boxes are the predicted labels from our VGQC w/ fusion, and the last row contains the corresponding queries.}

  \label{fig:qualitative-result}
\end{figure}

\paragraph{Qualitative results.}
To understand how VGQC performs better than TransVG in a qualitative manner, we show four examples from RefCOCO+ testA set in Figure~\ref{fig:qualitative-result}.
It is observed that when a query contains multiple objects, 
VGQC locates the target object correctly while TransVG fails. For example, ``female" and ``driver" are both mentioned in the query in Figure~\ref{fig:qualitative-result}(a). VGQC locates the correct target ``female", but TransVG mistakenly chooses ``driver".
In addition, QCM also outperforms TransVG in cases where there are multiple objects with different attributes. As shown in Figure~\ref{fig:qualitative-result}(b), (c), and (d), TransVG can correctly locate a person, 
but it fails to select the one matching the attributes in the queries while VGQC succeeds. This indicates that VGQC has a stronger ability to understand the image with respect to the query.

    


\begin{figure}[h]
  \centering
    
  \includegraphics[width=1\linewidth]{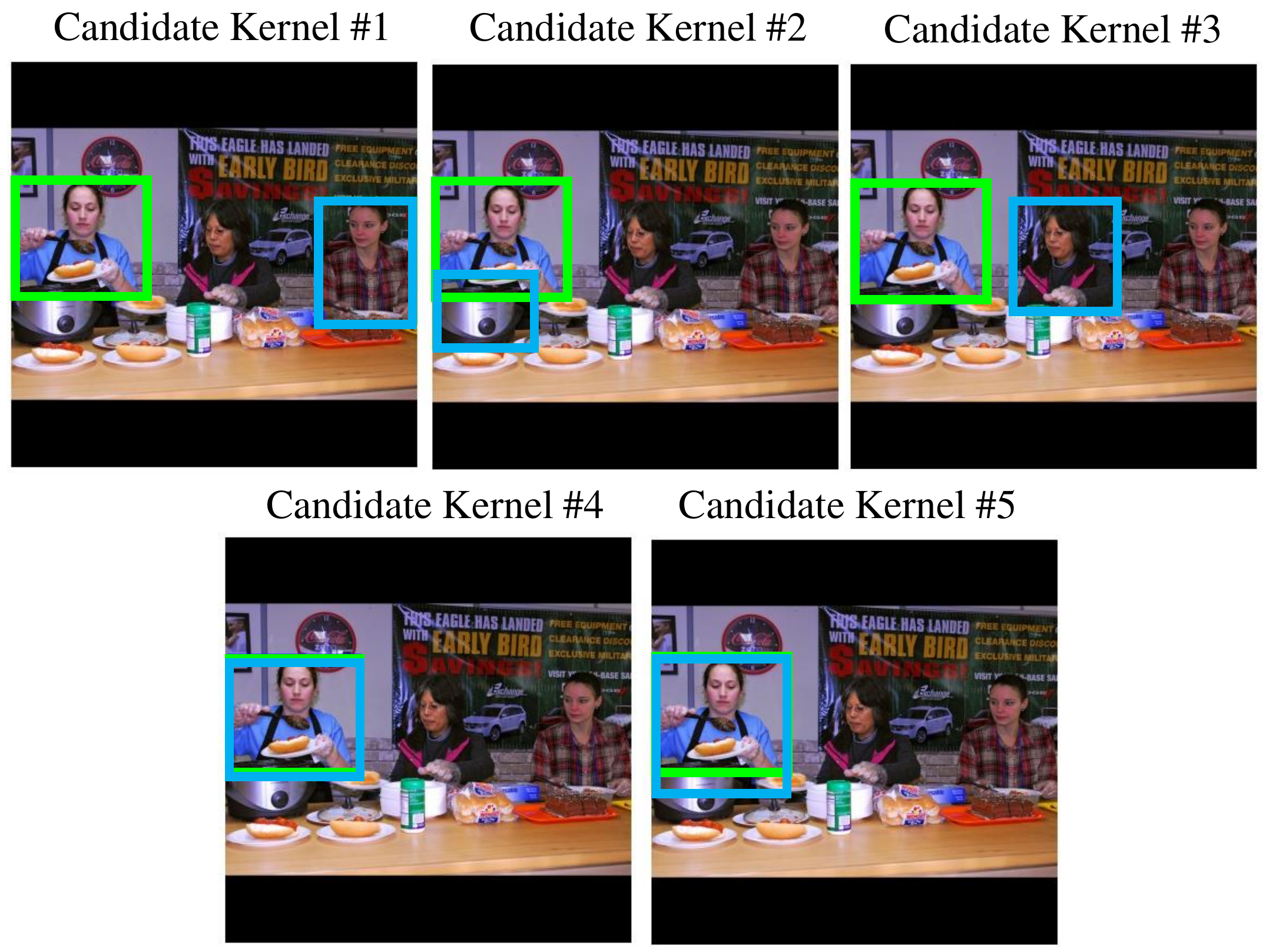}
  \caption{Ablations on candidate kernels. We force the model to rely on only one candidate kernel in the final QCM block and show the predictions. The \textcolor{green}{green} boxes are the ground truth labels and the \textcolor{cyan}{blue} boxes are the predicted labels from our VGQC w/ fusion. The query is ``woman serving the sub". Using different candidate kernels give different prediction results.}

  \label{fig:candidate-kernel-ablation}
\end{figure}
\paragraph{Diversity of QCM candidate kernels.}
To show that each candidate kernel encodes an image from a different perspective and has the potential to produce expressive aggregated kernels, we study the diversity of candidate kernels by examining VGQC's performance when it is forced to exclusively use one of the candidate kernels at the final QCM block.
As shown in Figure~\ref{fig:candidate-kernel-ablation}, VGQC predicts a different object when relying on each one of the five candidate kernels. 
Although two of the candidate kernels (\#4 and \#5) can give the correct prediction on their own with minor deviation, such one-candidate-kernel setting leads to a significant performance drop on the overall datasets shown in Table~\ref{tab:candidate-kernel-ablation}. Thus, we conclude that QCM uses a diverse set of candidate kernels to achieve competent performance.



\begin{figure}[h]
  \centering
  \includegraphics[width=1.0\linewidth]{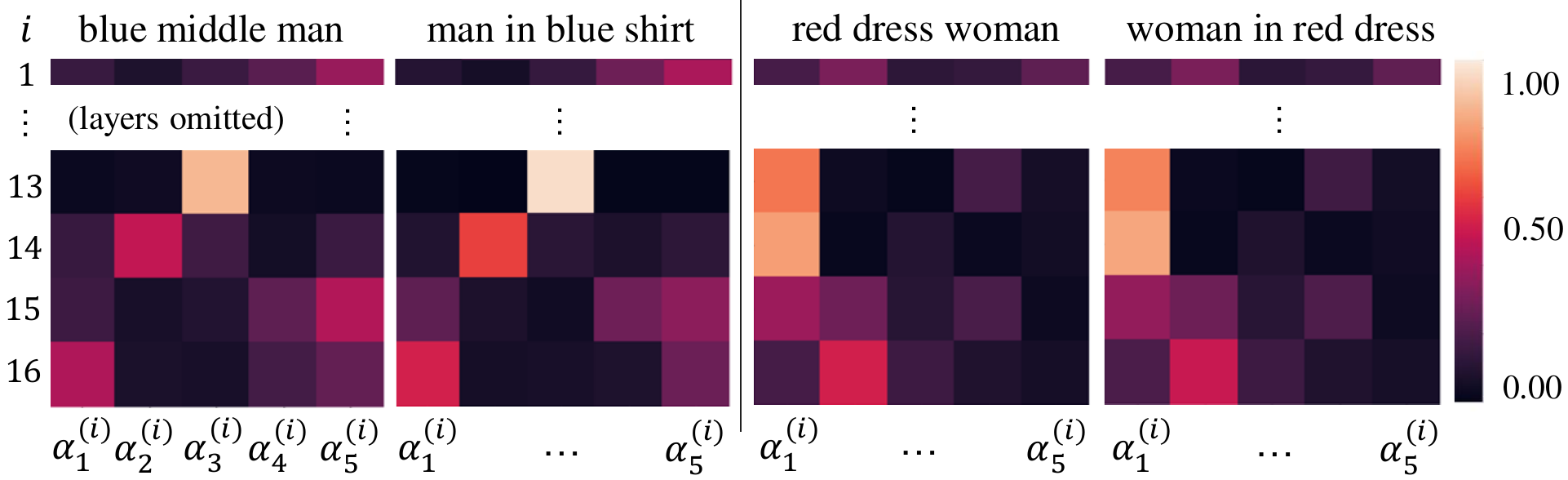}
  \caption{Attention weights of the last 4 out of 16 QCM blocks of VGQC w/ fusion model trained on RefCOCO, with 4 different queries shown at the top. $\alpha_k^{(i)}$ represents the attention weight for the $k$-th candidate kernel at block $i$.}
  \label{fig:blue-red-atten}
\end{figure}
\paragraph{Attention in QCM with query semantics.} By analyzing the attention weights in QCM blocks, we show that they carry the semantics of queries from query tokens. 
In Figure~\ref{fig:blue-red-atten}, we visualize the attention weights in the last 4 out of the total 16 QCM blocks of VGQC w/ fusion with heat maps. Each attention weight is a 5-dimensional vector. 
It can be observed that queries with similar semantics, e.g. ``blue middle man'' and ``man in blue shirt'', as well as ``red dress woman'' and ``woman in red dress'', have similar attention weights. Contrarily, queries with different semantics, e.g. ``blue middle man'' and ``red dress woman'', have dissimilar attention weights. 
This implies that QCM successfully incorporates 
the sentence representation into the attention weights, resulting in a meaningful weight assignment to the candidate kernels.
\begin{table*}[h]
\centering
\vspace{-0.3cm}
\begin{minipage}[t]{0.33\textwidth}
\makeatletter\def\@captype{table}
\centering
\scalebox{0.65}{
    \begin{tabular}{lrrrrrr}
    \toprule
          & All & \#1 & \#2 & \#3 & \#4 & \#5  \\ \hline
    val   & \textbf{69.17} & 54.80    &  43.54   & 58.73    & 57.55    & 55.76     \\
    testA & \textbf{73.52} & 57.28    &  30.73   & 58.66    & 58.69    & 58.87     \\
    testB & \textbf{58.97} & 49.96    &  55.83   & 54.48    & 57.55    & 49.94      \\ \hline
    \end{tabular}}
    \caption{Overall accuracy (\%) on RefCOCO+ when VGQC w/ fusion uses only one candidate kernel in the final QCM block. 
    Their performance is worse than the normal VGQC w/ fusion model (column ``All"). 
    }
    \label{tab:candidate-kernel-ablation}
\end{minipage}
\hspace{0.01\textwidth}
\begin{minipage}[t]{0.3\textwidth}
\makeatletter\def\@captype{table}
\centering
\scalebox{0.6}{
    \begin{tabular}{llrr}
    \toprule
    \multirow{2}{*}{Query1} & \multirow{2}{*}{Query2} & VGQC & VGQC \\
    &                       & w/ fusion                 & w/o fusion \\
    \midrule
    woman in red dress & lady red & 0.0001 & 0.0106\\ \hline
    woman in red dress & striped shirt & 0.5794 & 0.2779\\ \hline
    \makecell[l]{man in striped shirt \\on bike} & striped shirt & 0.0033 & 0.0025 \\ 
    \bottomrule
    \end{tabular}}
    \caption{Cosine distances of aggregated kernels for different query pairs in QCM block {\#10}. The model is trained on RefCOCO.}
    \label{tab:dynamic-kernel-eachother}
\end{minipage}
\hspace{0.01\textwidth}
\begin{minipage}[t]{0.3\textwidth}
\makeatletter\def\@captype{table}
\centering
\scalebox{0.65}{
    \begin{tabular}{lrrr}
    \toprule
    Models                 & val   & testA & testB \\
    \midrule
    TransVG                   & 80.18 & 82.53 & 75.25 \\ 
    VGQC w/ fusion (a) & 80.81 & 82.44 & 76.33 \\  
    VGQC w/ fusion (b) & 80.66 & 82.42 & 75.39 \\  
    VGQC w/ fusion            & \textbf{82.47} & \textbf{85.02} & \textbf{77.98} \\ 
    \bottomrule
    \end{tabular}}
    \caption{Comparing different derivations of QCM's attention weights $\boldsymbol{\alpha}$ on RefCOCO. (a): $\boldsymbol{\alpha}$ is a learnable vector on its own. (b): $\boldsymbol{\alpha}$ is derived from feature maps.}
    
    \label{tab:attention-ablation}
\end{minipage}
\end{table*}
\begin{table*}[h]
\vspace{-0.4cm}
\centering
\begin{minipage}[t]{0.33\textwidth}
\makeatletter\def\@captype{table}
\centering
\scalebox{0.65}{
    \begin{tabular}{lrrr}
    \toprule
    Structure                 & val   & testA & testB \\
    \midrule
    TransVG                   & \textbf{80.18} & 82.53 & \textbf{75.25} \\
    VGQC-layer 2\&3\&4  & 80.01 & \textbf{83.02} & 73.94 \\
    VGQC-layer 3\&4  & 79.11 & 81.70 & 72.23 \\
    VGQC-layer 4  & 78.93 & 82.37 & 73.08 \\
    \bottomrule
    \end{tabular}}
    \caption{Ablation study on VGQC w/o fusion model trained on RefCOCO where we apply QCM blocks only to the later layers of the visual encoder backbone.}
    \label{tab:layer-ablation}
\end{minipage}
\hspace{0.01\textwidth}
\begin{minipage}[t]{0.3\textwidth}
\makeatletter\def\@captype{table}
\centering
\scalebox{0.65}{
    \begin{tabular}{rrrr}
    \toprule
    K & val & testA & testB \\
    \midrule
    3 & 79.24 & \textbf{83.04} & 73.69 \\
    5 & \textbf{80.01} & 83.02 & \textbf{73.94} \\
    8 & 79.05 & 81.96 & 73.23 \\
    \bottomrule
    \end{tabular}}
    \caption{Ablation study on different number of candidate kernels (K) in QCM block based on VGQC w/o fusion trained on RefCOCO dataset.}
    \label{tab:num-kernel-ablation}
\end{minipage}
\hspace{0.01\textwidth}
\begin{minipage}[t]{0.3\textwidth}
\makeatletter\def\@captype{table}
\centering
\scalebox{0.65}{
    \begin{tabular}{lrrr}
    \toprule
    Model & val & testA & testB \\
    \midrule
    TransVG-6-layers        & 80.18 & 82.53 & 75.25 \\
    VGQC w/ fusion-6-layers  & 82.47 & \textbf{85.02} & 77.98 \\
    VGQC w/ fusion-4-layers  & \textbf{82.67} & 84.54 & \textbf{78.26} \\
    VGQC w/ fusion-2-layers  & 81.61 & 84.68 & 76.57 \\
    VGQC w/o fusion         & 80.01 & 83.02 & 73.94 \\
    \bottomrule
    \end{tabular}}
    \caption{Ablation study on the number of layers used in the fusion transformer evaluated on RefCOCO dataset.}
    \label{tab:num-fusion-layer-ablation}
\end{minipage}
\vspace{-0.5cm}
\end{table*}
\paragraph{Aggregated kernels with query semantics.}
The query semantics carried by attention weights are further propagated to aggregated kernels. 
Specifically, kernels composed from similar text queries have a closer cosine distance between each other in their own space. 
We pass in different queries to VGQC and take the aggregated kernels from QCM block {\#10}. The aggregated kernels are flattened into vectors to calculate the cosine distances, and their distances between each other are shown in Table~\ref{tab:dynamic-kernel-eachother}. 
The aggregated kernels from ``woman in red dress” and ``lady red” are close to each other, and the distance between the aggregated kernels from ``man in striped shirt on bike” and ``striped shirt” are also small. On the contrary, ``woman in red dress” and ``striped shirt” have different semantics and their aggregated kernels are far apart from each other.
This demonstrates that the aggregated convolution kernels are embedded with meaningful textual information to extract query-aware visual features. 

\subsection{Ablation Study}
\paragraph{Convolution kernel generation from text and image.}
We show that the performance improvement brought by QCM is not attributed to the introduction of new parameters (the attention and candidate kernels), but the incorporation of textual information. In our experiments, we apply two other methods to derive the attention weight vector $\boldsymbol{\alpha}$ and thus the aggregated kernel in every QCM block. Method (a): $\boldsymbol{\alpha}$ is randomly initialized as a learnable $K$-dimensional vector. Method (b): $\boldsymbol{\alpha}$ is computed from the output feature map of the previous CNN layer through average-pooling and a linear projection followed by a softmax layer. We compare them with our main method, where the attention weights are computed from the text query, as illustrated in Section~\ref{sec:visual-encoder}. All three methods utilize attention weights to compose the aggregated kernels, but (a) and (b) do not incorporate textual information during visual feature extraction.
The results in Table~\ref{tab:attention-ablation} on RefCOCO dataset show that incorporating textual information results in the largest performance improvement, and
adding extra parameters without leveraging textual information brings only insignificant improvements. This demonstrates the effectiveness of QCM.

\paragraph{Number of query-conditioned convolution blocks.} 
We study the impact of start applying QCM at different layers in the visual encoder.
In the ResNet backbone which has 4 layers of convolution blocks, we use QCM in layer 4, layer $3\&4$, and layer $2\&3\&$4. The experiment is conducted on VGQC w/o fusion on RefCOCO dataset. The result in Table~\ref{tab:layer-ablation} shows that the performance drops when applying QCM to only later layers,
which means it is beneficial to start incorporating textual information from earlier layers.

\paragraph{Number of candidate kernels.} We measure the effect of changing the number of candidate kernels used in our QCM blocks. We trained three VGQC w/o fusion models on RefCOCO dataset and evaluate them in all three splits. Among the three settings in Table~\ref{tab:num-kernel-ablation}, the model with 5 candidate kernels performs the best, and having fewer or more candidate kernels negatively affects the performance.

\paragraph{Number of fusion layers.} We show that query-aware visual representations produced by visual encoders with QCM are better in the sense that it requires less or no further fusion layers. As shown in Table~\ref{tab:num-fusion-layer-ablation}, our method outperforms TransVG with fewer fusion layers, and VGQC with a 4-layer fusion module has comparable performance to VGQC with a 6-layer fusion module.

\section{Conclusion}
In this paper, we propose a query-conditioned convolution module (QCM) to be integrated into visual encoders to extract query-aware visual features. We also present a query-conditioned visual grounding pipeline (VGQC) utilizing QCM to address the limitation of the previous extract-and-fuse pipeline, where the extracted visual features may not convey the required information described in the text query.
Our query-aware visual features not only facilitate the multi-modal interaction but are also informative enough to be directly used for prediction without further multi-modal fusion. Experiments show that VGQC w/ fusion outperforms all previous one-stage visual grounding methods and achieves state-of-the-art performance, and VGQC w/o fusion increases inference speed with a simpler structure while reaching comparable performance as the latest methods. In addition, the extensive analysis and visualizations show that QCM can effectively comprehend query information and extract query-aware visual features. Finally, our method is applicable to other textual-visual tasks, which merits further investigation in future work.



\newpage
\bibliographystyle{named}
\bibliography{ijcai22}

\end{document}